\documentclass[sigconf]{acmart}
\AtBeginDocument{%
  \providecommand\BibTeX{{%
    \normalfont B\kern-0.5em{\scshape i\kern-0.25em b}\kern-0.8em\TeX}}}

\copyrightyear{2022} 
\acmYear{2022}
\acmDOI{10.1145/XXXX.XXXX}
\setcopyright{rightsretained}
\acmConference[CIKM '22]{Proceedings of the 31st ACM International Conference on Information and Knowledge Management}{October 17--21, 2022}{Atlanta, GA, USA}
\acmBooktitle{Proceedings of the 31st ACM International Conference on Information and Knowledge Management (CIKM '22), October 17--21, 2022, Atlanta, GA, USA}
\acmDOI{10.1145/3511808.3557420} \acmISBN{978-1-4503-9236-5/22/10}

\usepackage[linesnumbered,ruled,vlined,algo2e]{algorithm2e}
\usepackage{multirow}
\usepackage{tabularx}
\usepackage{graphicx}
\usepackage{float}
\usepackage{subfigure}
\usepackage{array}
\usepackage{amsfonts}
\usepackage{epsfig}
\usepackage{graphicx}
\usepackage{amsmath}
\usepackage{booktabs}
\usepackage{appendix}
\usepackage{bbm}
\usepackage{stfloats}
\usepackage{flushend}

\usepackage{amsmath,amsfonts,amsthm}
\usepackage{mathtools}
\usepackage{commath}
\usepackage{hyperref}

\newcommand{\PreserveBackslash}[1]{\let\temp=\\#1\let\\=\temp}
\newcolumntype{C}[1]{>{\PreserveBackslash\centering}p{#1}}
\newcolumntype{R}[1]{>{\PreserveBackslash\raggedleft}p{#1}}
\newcolumntype{L}[1]{>{\PreserveBackslash\raggedright}p{#1}}

\newcommand{\ours}{hiNet\xspace}
\newcommand{\our}{hiNet}

\newcommand{\spo}{SpO\textsubscript{2}\xspace}
\newcommand{\Vital}{\mathcal{V}\xspace}
\newcommand{\vital}{\mathbf{v}\xspace}

\newcommand{\MTS}{\mathbf{X}\xspace}
\newcommand{\mts}{\mathbf{x}\xspace}
\newcommand{\reMTS}{\hat{\mathbf{X}}\xspace}
\newcommand{\remts}{\hat{\mathbf{x}}\xspace}
\newcommand{\win}{W\xspace}
\newcommand{\rep}{\mathbf{z}}

\newcommand{\nop}[1]{}

\newtheorem{problem}{Problem}

\usepackage{enumitem}
\setlist[itemize]{leftmargin=*}

\begin{abstract}
We present an end-to-end model using streaming physiological time series to predict near-term risk for hypoxemia, a rare, but life-threatening condition known to cause serious patient harm during surgery. Inspired by the fact that a hypoxemia event is defined based on a future sequence of low \spo (i.e., blood oxygen saturation) instances, we propose the hybrid inference network (hiNet) that makes hybrid inference on both future low \spo instances and hypoxemia outcomes. hiNet integrates 1) a joint sequence autoencoder that simultaneously optimizes a discriminative decoder for label prediction, and 2) two auxiliary decoders trained for data reconstruction and forecast, which seamlessly learn contextual latent representations that capture the transition from present states to future states. All decoders share a memory-based encoder that helps capture the global dynamics of patient measurement. For a large surgical cohort of 72,081 surgeries at a major academic medical center, our model outperforms strong baselines including the model used by the state-of-the-art hypoxemia prediction system. With its capability to make real-time predictions of near-term hypoxemic at clinically acceptable alarm rates, hiNet shows promise in improving clinical decision making and easing burden of perioperative care.

\end{abstract}



\begin{document}
\title{Predicting Intraoperative Hypoxemia with Hybrid Inference Sequence Autoencoder Networks}


\author{Hanyang Liu}
\affiliation{%
  \institution{McKelvey School of Engineering \\ Washington University in St. Louis}
  \streetaddress{1 Brooking Drive}
  \city{St. Louis}
  \country{United States}
  \postcode{63130}
}
\email{hanyang.liu@wustl.edu}

\author{Michael C. Montana}
\affiliation{%
  \institution{School of Medicine \\ Washington University in St. Louis}
  \city{St. Louis}
  \country{United States}
  \postcode{63130}
}
\email{montana@wustl.edu}

\author{Dingwen Li}
\affiliation{%
  \institution{McKelvey School of Engineering \\ Washington University in St. Louis}
  \streetaddress{1 Brooking Drive}
  \city{St. Louis}
  \country{United States}
}
\email{dingwenli@wustl.edu}

\author{Chase Renfroe}
\affiliation{%
  \institution{School of Medicine \\ Washington University in St. Louis}
  \city{St. Louis}
  \country{United States}
}
\email{stephen.renfroe@wustl.edu}

\author{Thomas Kannampallil}
\affiliation{%
  \institution{School of Medicine \\ Washington University in St. Louis}
  \city{St. Louis}
  \country{United States}
}
\email{thomas.k@wustl.edu}

\author{Chenyang Lu}
\authornote{Corresponding author.}
\affiliation{%
  \institution{McKelvey School of Engineering \\ Washington University in St. Louis}
  \streetaddress{1 Brooking Drive}
  \city{St. Louis}
  \country{United States}
}
\email{lu@wustl.edu}

\renewcommand{\shortauthors}{Hanyang Liu et al.}

\maketitle
\section{Introduction}
Hypoxemia, or low blood oxygen saturation (\spo), is an adverse physiological condition known to cause serious patient harm during surgery and general anaesthesia~\cite{dunham2014perioperative}. Without early intervention, prolonged hypoxemia can seriously affect surgical outcomes and is associated with many adverse complications such as cardiac arrest,  encephalopathy, delirium, and post-operative infections~\cite{lundberg2018explainable}. 
To mitigate the effects of hypoxemia, anesthesiologists monitor \spo levels during general anesthesia using pulse oximetry, so that actions can be taken in a timely manner. Despite the availability of real-time data, however, reliable prediction of hypoxemic events remains elusive ~\cite{ehrenfeld2010incidence}.

In recent years, data from electronic health records (EHR) have been used to develop predictive models to anticipate risks of future adverse events, facilitating early interventions to mitigate the occurrence of adverse events~\cite{west2016interventions,liu2022hipal,lou2022predicting}. Similar attempts~\cite{elmoaget2016multi,erion2017anesthesiologist} have been made to target hypoxemia based on \spo data. Recently, \cite{lundberg2018explainable} proposed a gradient boosting machine (GBM) model for predicting intraoperative hypoxemia by integrating a number of preoperative static variables and intraoperative physiological time series. Compared to prior works that only utilized \spo, the use of multi-modal data helps train a more reliable prediction model. However, as classical models such as GBM cannot directly utilize multivariate time series, they require the extraction of hand-crafted features with limited capacity for capturing the temporal dynamics of time series. In addition, the uniquely selected combination of features may not generalize to a new cohort from another hospital.

Another major limitation of the aforementioned hypoxemia models is that they were developed to target any low \spo occurrences where most of them are short-term (e.g., a single minute) and transient \spo drop. In practice, most occurrences of low \spo do not necessarily reflect high patient risks for deterioration that actually require anesthesiologist intervention~\cite{laffin2020severity}. In contrast to transient \spo reduction, significant risks arise from \textit{persistent hypoxemia}, defined as continuously low \spo over a longer time window (e.g., 5 minutes). Persistent hypoxemia can develop rapidly and unexpectedly due to acute respiratory failure, or other circumstances. It is immediately life-threatening if not treated ~\cite{mehta2016management}.
For this study we have collected a large surgical dataset from a major academic medical center. In this cohort, despite the rareness (1.5\% of all time) of low \spo, 24.0\% of the surgical encounters experienced at least a single instance of \spo drop (measured per minute), while merely 1.9\% experienced persistent hypoxemia (over a 5-minute time window). Unessential alarms can result in clinicians' desensitization to alarms, and thus the actually critical ones could possibly be missed. Hence, it is of high clinical importance to reliably predict \textit{persistent} hypoxemia (i.e., with high sensitivity of predicting advert events with clinically acceptable alarm rate), which facilitates the most pivotal early interventions. However, the prediction of persistent hypoxemia is a greater challenge, as it is difficult for a machine learning model to learn reliable patterns from data with a certain class being underrepresented (0.13\% positive rate in our real-world dataset). Moreover, it requires the model to foresee the future over a longer horizon, when the past data are decreasingly indicative of distal future outcomes.

We aim to address these challenges by developing an end-to-end learning framework that utilizes streaming physiological
time series (e.g., heart rate, \spo) and produces risk prediction of hypoxemia, while simultaneously learning powerful latent representation known to improve model robustness against class imbalance~\cite{hendrycks2019using}. In addition to general hypoxemia (i.e., any \spo drop), we focus on predicting persistent hypoxemia given its clinical significance. Intuitively, if we can forecast future input data, especially the \spo variation, we can anticipate the potential hypoxemia risk more accurately. We propose a novel deep model, the \underline{h}ybrid \underline{i}nference \underline{net}work (\ours) that simultaneously
makes inference on both future hypoxemic events and the sequence of future \spo levels. This end-to-end framework is enabled by jointly optimizing: (\textbf{\textit{i}}) a memory-augmented sequence encoder that both aggregate local temporal features and capture global patient dynamics; (\textbf{\textit{ii}}) a sequence decoder for data reconstruction; (\textbf{\textit{iii}}) a sequence decoder that models the evolution of future \spo levels; and (\textbf{\textit{iv}}) a discriminative decoder (classifier) trained for hypoxemia event prediction. With joint training, the classifier can leverage the learned latent representation, while the supervisory signal from the classifier can be propagated to seamlessly direct representation learning towards optimizing the desired prediction task.

The proposed model is trained and evaluated on a large real-world pediatric cohort from a large academic medical center. The data includes minute-resolution multi-modal time series collected from 72,018 surgeries that correspond to 118,000 hours of surgery. The experiments show that our proposed model can reliably predict both general and persistent hypoxemia more precisely with lower alarm rates than a set of strong baselines including the model employed by the state-of-the-art hypoxemia prediction system.

Specifically, our contributions are threefold:
\begin{itemize}
\vspace{-1em}
\item We propose the first learning-based approach for \textit{persistent} hypoxemia prediction, a challenging but clinically significant problem.

\item We design a novel sequence learning framework for multivariate time series that jointly optimizes multiple highly-correlated tasks including a supervised discriminative task and two sequence generation tasks. 
Through joint training, the learned contextual latent representations facilitate better predictions and meanwhile are seamlessly optimized for task-specific effectiveness.

\item Extensive experiments on a large pediatric surgical cohort show the improvement of our proposed model over strong baselines and the potential of hiNet  to support clinical decisions and impact surgical practice.

\end{itemize}

\section{Hypoxemia Prediction Problem}\label{sec:formulation}

\begin{figure*}[t]
  \centering
  \includegraphics[width=0.95\textwidth]{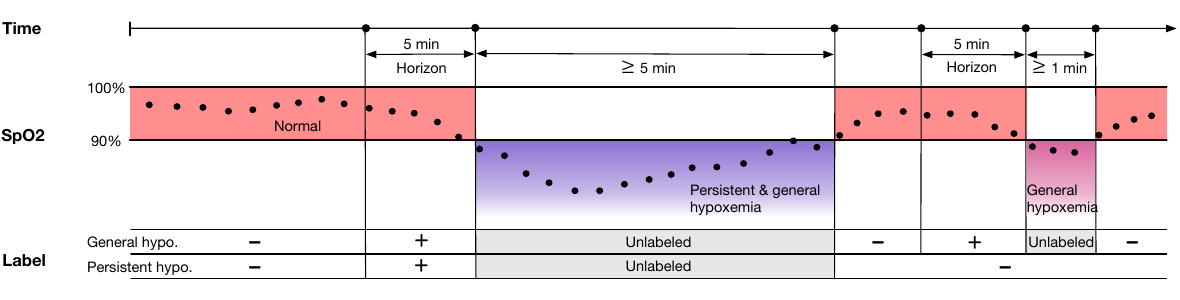}
  \captionof{figure}{Definition of hypoxemia in two severeness levels and the training label assignment.}
  \label{fig:labels}
\end{figure*}

\subsection{Intraoperative Time Series Data}\label{sec:data}
During anesthesia procedures (i.e., surgeries), a set of patient's physiological signals such as vital signs (e.g., heart rate, \spo, $\text{CO}_2$) and ventilator parameters (e.g., respiration rate, $\text{FiO}_2$) are recorded at one-minute intervals. These intraoperative time series track the patient's physical status during surgery and may contain information associated with potential complications and adverse surgical events~\cite{meyer2018machine}.

The data used in this work were collected from 79,142 pediatric surgical encounters spanning from 2014 to 2018 with approximately 118,000 hours of surgeries in total (89 min per case in average) at the St. Louis Children’s Hospital, a free-standing tertiary care pediatric hospital, and St. Louis Children’s Specialty Care Center, an outpatient pediatric surgical center. The institutional review board of Washington University approved this study with a waiver of consent (IRB \#201906030).

\subsection{Hypoxemia Definition and Labeling}\label{sec:hypo_def}

We use a stringent clinical definition of hypoxemia to assess a patient encounter during surgery. We follow the guideline recommended by World Health Organization \cite{WHO2011pulse} and use the emergency level of \spo ($\leq 90\%$) as the threshold for hypoxemia.
We define two types of hypoxemia events based on two different severeness levels and assign labels for the two prediction problems following the criteria as shown in Figure 3. 
\begin{itemize}
    \item \textbf{General hypoxemia}: Any low SpO2 ($\leq 90\%$) instance (similar to prior work \cite{lundberg2018explainable}, light purple region).
    \item \textbf{Persistent hypoxemia}: Low \spo ($\leq 90\%$) instance that consecutively lasts for 5 minutes and more (dark purple region).
\end{itemize}
In clinical practice, temporary drops in \spo (i.e., general hypoxemia) are common (24.0\% patient encounters in our dataset) and less concerning which is usually rapidly correctable with simple maneuvers with no short-or long-term sequelae. Persistent hypoxemia (1.9\% patient encounters) is more clinically relevant and can rapidly become life-threatening and also much more difficult to predict since we need to anticipate deeper into the future.

Figure 3 shows how labels are assigned for model development. For each prediction problems, given a 5-minute prediction horizon, we assign positive labels to the time steps within this 5 min predictive window right before the start of a hypoxemic event (persistent or general hypoxemia, respectively). Other time steps prior to this predictive window are assigned with negative labels. We leave the samples during the time window when the patient is already in a hypoxemic event unlabeled, as it has little clinical benefit to predict hypoxemia when hypoxemia already occurs.

\subsection{Near-term Prediction Problem}

Our dataset consists of $N$ independent surgeries, denoted as $\mathcal{D}=\{\mathcal{V}_i\}_{i=1}^{N}$, where $i$ is the index of surgeries, and $\Vital_i$ is a set of time series inputs. We assume the time span is divided into equal-length time intervals. The multi-channel time series $\Vital_i=[\vital_{i}^1, \vital_{i}^2, ..., \vital_{i}^{T_i}]\in\mathbb{R}^{V\times T_i}$, where $V$ is the number of time series channels, $T_i$ is the length of surgery $i$, and $\vital_{i}^t\in\mathbb{R}^{V}$ is the vector of recorded data values at timestep $t$. For timestep $t$, we have a binary label $y_{i,t}\in\{0,1\}$, where $1$ indicates that a hypoxemic event will occur anytime within the next fixed-length time window $[t+1, t+\win_h]$, otherwise $y_{i, t}=0$. $\win_h$ denotes the prediction horizon. We aim to solve the following:
\begin{problem}
\em{
Given a new surgery $i$ where the patient is not already in hypoxemia, and the data window $\MTS_{i,t} = \Vital_i^{(t-\win_o,t]} \in\mathbb{R}^{V \times \win_o}$ at time $t$ (zero padded if $1\leq t<\win_o$),
the goal is to train a classifier $f$ that produces the label $y_{i,t}$: $y_{i,t} = f(\MTS_{i,t})$.}
\end{problem}

\section{Related Works}

\subsection{Learning to Predict Hypoxemia}
Recently, several attempts have been made targeting hypoxemia using data-driven approaches. For instance, \cite{elmoaget2016multi} used a linear auto-regressive (AR) model on \spo forecast. \cite{erion2017anesthesiologist} sought to use deep learning models such as LSTM to directly classify past \spo data. \cite{nguyen2018reducing} used AdaBoost to identify false \spo alarms, without directly targeting prediction. Recently, a more comprehensive approach, Prescience~\cite{lundberg2018explainable}, employed GBM to predict general hypoxemia based on both preoperative static variables and intraoperative time series. All these approaches aimed at either forecasting \spo (regression) or predicting only general hypoxemia.  


\begin{figure}[t]
    \centering
    \includegraphics[width=0.99\linewidth]{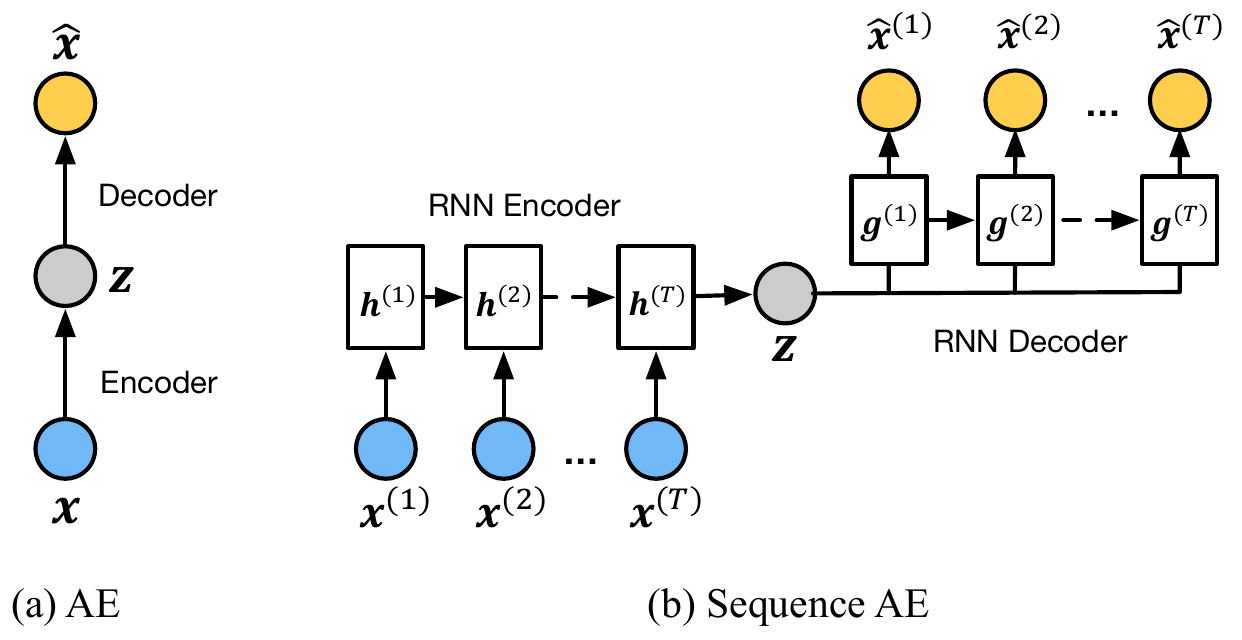}
    \caption{Generalized AE and sequence AE. In sequence AEs, usually an RNN (e.g., LSTM) is used for both aggregation and disaggregation of sequential data.}
    \label{fig:lstm_ae}
\end{figure}

\subsection{Encoder-decoder Sequential Learning}


Autoencoder (AE)~\cite{davis2006relationship}, as shown in \autoref{fig:lstm_ae}(a), is widely used for representation learning. AE is trained to simultaneously optimize an \textit{encoder} that maps input into latent representation, and a \textit{decoder} that recovers the input by minimizing the reconstruction error.
However, AE cannot directly handle sequential data. Recently, \cite{dai2015semi} proposed a \textit{seq2seq} AE by instantiating the encoder and decoder with LSTM, referred to as LSTM-AE, shown in \autoref{fig:lstm_ae}(b). \cite{srivastava2015unsupervised} further extended it to the composite LSTM autoencoder (LSTM-CAE) that additionally trains another decoder fitting future data as regularization. The LSTM-AE based methods have shown promising performance in learning representation for sequential data. For instance, \cite{laptev2017time,zhu2017deep} proposed to use pretrained LSTM-AE to extract deep features from time series for Uber trips and rare event forecasting. Recent clinical applications can be found in \cite{suresh2017use,ballinger2018deepheart,baytas2017patient} that used LSTM-AE to extract patient-level representation for phenotyping and cardiovascular risk prediction. These works are related to our approach, as they all use representation learning of pretrained LSTM-AE to facilitate a classification task. 
However, with the goal to continuously provide real-time prediction, instead of extracting patient-level representation, we use a sequence AE to aggregate local data sequence and learning representation of a data window sliding on each surgical trajectory. 
Unsupervised pretraining tends to learn general task-agnostic underlying structure, and thus the greedy layer-wise optimization in separate steps can lead to a suboptimal classifier~\cite{zhou2014joint}. Instead, our approach relies on building an end-to-end model that jointly optimizes the classification and latent representation learning while balancing them more delicately.



\section{The HiNet Framework}
\begin{figure*}[t]
    \centering
    \includegraphics[width=0.99\textwidth]{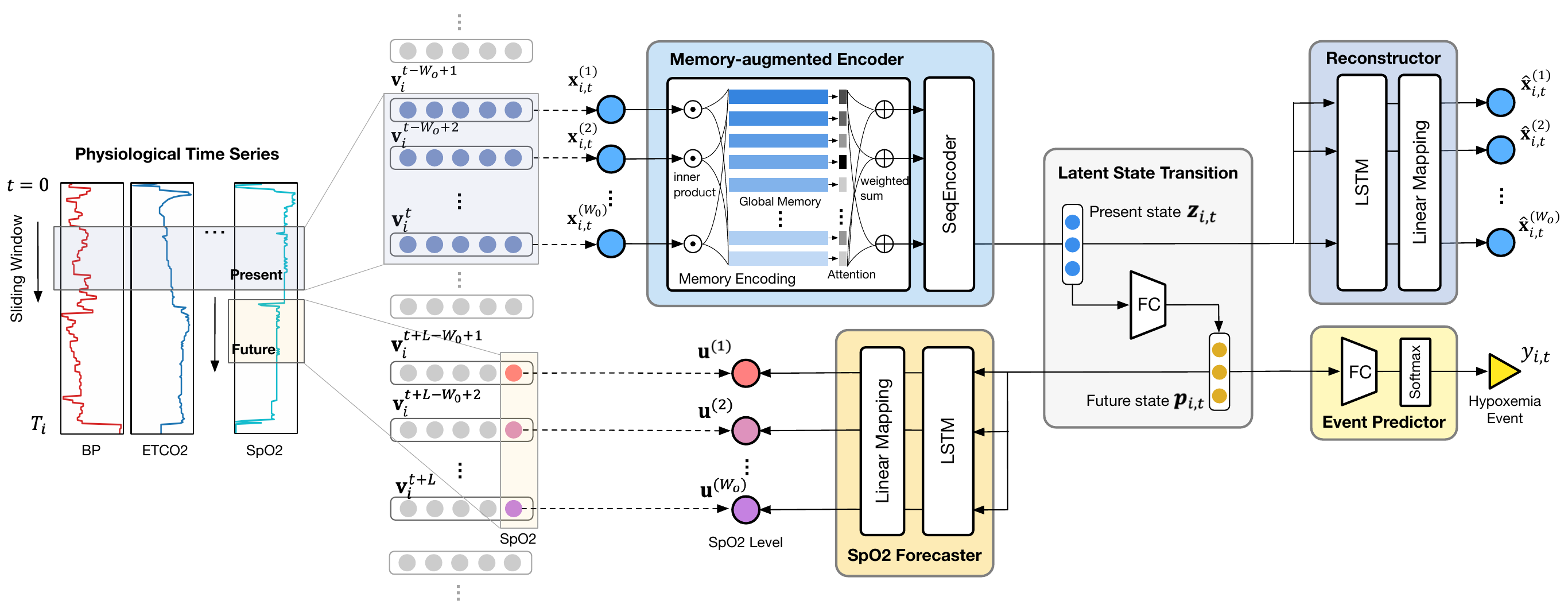}
    \caption{The end-to-end architecture of \ours. The streaming data are fed into the memory-based encoder. The encoder-decoder network jointly optimizes \textbf{Reconstructor} trained to recover input, \textbf{Forecaster} trained to model the transition of future low \spo instances, and \textbf{Event Predictor} to estimate event labels. 
    }
    \label{fig:framework}
\end{figure*}
Figure \ref{fig:framework} shows an overview of our approach. This end-to-end framework jointly optimizes the desired classification task for prediction, and two auxiliary tasks for representation learning. The addition of the sequence forecasting decoder contributes to learning future-related contextual representation. The joint training allows the supervised loss to direct representation learning towards being effective to the desired classification task. 
Hence, the hybrid integration of the three decoders enables the model to balance extracting the underlying structure of data and providing accurate prediction. 

\subsection{Memory-augmented Sequence Autoencoder}

When applying a sequence AE on streaming time series using a sliding data window, the sequence encoder tends to learn mainly the local temporal patterns within the window.
Recently, memory network~\cite{sukhbaatar2015end,gong2019memorizing} has shown promising results in data representation. Generally, a memory network updates an external memory consisting of a set of basis features for look-up to preserve general patterns optimized on the whole dataset~\cite{santoro2016meta}. 
To help capture the global dynamics of physiological time series, we design novel a memory augmented sequence AE with the dual-level embedding of the input sequences. 

\subsubsection{Dual-level Sequence Embedding}

Given a sliding window $\MTS_{i,t}=[\mts_{i,t}^{(k)}]_{k=1}^{\win_o}=[\vital_i^{t-\win_o+1}, ..., \vital_i^{t-1}, \vital_i^{t}] $, we first represent the features at each step using a set of basis vectors that memorize the most representative \textit{global} patterns across all surgery cases, and then use a sequence encoder to aggregate all the memory-encoded vectors within the \textit{local} input window into one vector as representation, as shown in Figure \ref{fig:framework}. 

\textit{Level 1: Step-level Global Memory Encoding.}
We assume there are $M$ feature basis vectors ($M$ as a hyperparameter) for a specific dataset, and initialize a \textit{global} memory $\mathbf{B}=[\mathbf{b}_1, \mathbf{b_2},...,\mathbf{b}_M]\in\mathbb{R}^{M\times V}$. We assume that features of each step can all be embedded to a linear combination of the $M$ feature bases. Given a feature vector $\mts^{(k)}$ of the $k$-th step, we obtain the attention $\alpha$ for each basis $\mathbf{b}_j$ by calculating the similarity of $\mts^{(k)}$ to each of them. The attention values are normalized by a softmax function: $\mathrm{Softmax}(z)=e^{z_k}/\sum_{k'}e^{z_{k'}}$. Then the embedded vector is the sum of all the bases weighted by their attention. The memory $\mathbf{B}$ is updated jointly with the network. More concretely,
\begin{equation}\label{eq:mem}
    \begin{split}
    \boldsymbol{\alpha}_{i,t}^{(k)} &= \mathrm{Softmax}(\mathbf{B}\mts_{i,t}^{(k)}) \\
    \mathbf{a}_{i,t}^{(k)} &= \sum_{j=1}^{M} \alpha_{i,t,j}^{(k)}\mathbf{b}_j
    \end{split}
\end{equation}

\textit{Level 2: Window-level Local Feature Aggregation.}
Now we use a standard sequence model parameterized by $\boldsymbol{\theta}_h$ to encode the \textit{local} temporal patterns within the embedded sequence. To make the framework more generalizable, here the sequence encoder $\Phi$ can be instantiated by any standard sequence model (e.g., LSTM, TCN \cite{bai2018empirical}, FCN \cite{wang2017time}). We use the output $\rep$ as the current representation for time $t$.
\begin{equation}\label{eq:lstm_e}
        \rep_{i,t}= \Phi([\mathbf{a}_{i,t}^{(1)}, \mathbf{a}_{i,t}^{(2)}, ..., \mathbf{a}_{i,t}^{(\win_o)}]; \boldsymbol{\theta}_h)
\end{equation}
Please see Appendix \ref{sec:design} for more details about the design choice of the sequence encoder base model.

For convenience, we denote this mapping from the input to the latent space $ p(\rep|\MTS_{i,t};\boldsymbol{\theta}_E):\mathbb{R}^{V \times \win_o} \xrightarrow{} \mathbb{R}^{Z}$ as \textit{Encoder}.
with trainable parameters $\boldsymbol{\theta}_E = \{\mathbf{B}, \boldsymbol{\theta}_h\}$. 


\subsubsection{Data Reconstruction}

We copy the vector $\mathbf{z}$ that represents the patient state at present for every step in the window $k\in [1, \win_o]$ as the input to the sequence disaggregation layers (i.e., an LSTM and then a linear layer as a surrogate of inverse memory) to reconstruct $\reMTS=[\hat{\mts}^{(1)}, \hat{\mts}^{(2)}, ..., \hat{\mts}_{(\win_o)}]$:
\begin{eqnarray}\label{eq:lstm_d}
        \mathbf{g}_{i,t}^{(k)} &=& \mathrm{LSTM}(\mathbf{z}_{i,t}, \mathbf{g}_{i,t}^{(k-1)}; \boldsymbol{\omega}_g) \\
        \label{eq:out_r}
        \remts_{i,t}^{(k)} &=&\mathbf{g}_{i,t}^{(k)} \mathbf{W}_g + \mathbf{d}_g
\end{eqnarray}
This mapping from the input space back to itself $f_R = p(\reMTS_{i,t}|\rep;\boldsymbol{\omega}_R):\mathbb{R}^{Z} \xrightarrow{} \mathbb{R}^{V \times \win_o}$ is denoted as as \textit{Reconstructor},
in which $\boldsymbol{\omega}_R=\{\boldsymbol{\omega}_q, \boldsymbol{\omega}_g, \mathbf{W}_g, \mathbf{d}_g\}$ represents the parameters of the sequence disaggregation layers. They can be learned by minimizing the loss: $\mathcal{L}_R = \mathrm{MSE}(\MTS_{i,t}, \reMTS_{i,t})$.

\subsection{Multi-decoder Hybrid Inference}
Reconstructor helps learn latent representation that improves model robustness against class imbalance. However, Reconstructor may not provide enough future-indicative patterns that the prediction task relies on.
Motivated to learn more future-related contextual representation, we propose the \underline{h}ybrid \underline{i}nference \underline{Net}work ({hiNet}) that incorporates both generative and discriminative components, and simultaneously makes inference to both sequence of future low \spo instances and hypoxemia event outcome. The overall architecture of \ours is shown in Figure \ref{fig:framework}.

\subsubsection{Latent State Transition}
Given the present patient state $\rep$, we use a fully-connected (FC) network parameterized by $\boldsymbol{\theta}_p$ to model the contextual transition to the patient state of the future $p$ in a time horizon $L$:
\begin{eqnarray}
    \mathbf{p}_{i,t} = \mathrm{FC}(\rep_{i,t}; \boldsymbol{\theta}_p) \label{eq:map_f}
\end{eqnarray}
This vector $\mathbf{p}$ that represents future patient state will be shared as the encoded data representation used by both a sequence forecasting decoder and a hypoxemia classifier, so that the future-indicative representation learning and the classification can seamlessly benefit from each other.

\subsubsection{Future \spo Forecast}
Since future hypoxemia events are strictly defined based on the sequence of future \spo levels (i.e., whether \spo$\leq 90\%$ and how long it lasts), we build another sequence decoder to forecast the future \spo sequence with a time horizon $L$, using the future state representation $\mathbf{p}$. We use ${\mathbf{u}}=[{u}^{(1)}, {u}^{(2)}, ..., {u}^{(\win_o)}]\in\mathbb{B}$ that corresponds to the future data window $[\vital_i^{t-\win_o+1+L}, ..., \vital_i^{t+L}]$ to denote \spo levels.
Similar to Eq. (\ref{eq:lstm_d}) and (\ref{eq:out_r}), we apply sequence disaggregation layers to the copied vectors $\mathbf{p}$.
We denote the mapping from input to future input, $f_F:\mathbb{R}^{V \times \win_o} \xrightarrow{} \mathbb{R}^{V \times \win_o}$ as \textit{Forecaster} . Hence
\begin{equation}\label{eq:Forecaster}
    \hat{\mathbf{u}}_{i,t+L} = f_F(\MTS_{i,t}; \boldsymbol{\theta}_E, \boldsymbol{\theta}_p, \boldsymbol{\omega}_F)
\end{equation}
where $\boldsymbol{\omega}_F$ denotes the task-specific parameters in the sequence disaggregation layers.
The parameters can be learned by minimizing the loss: $\mathcal{L}_R = \mathrm{MSE}(\MTS_{i,t+L}, \reMTS_{i,t+L})$.

\subsubsection{Hypoxemia Event Prediction}
Given the new representation $\mathbf{p}$ from Eq. (\ref{eq:map_f}) that contains indicative patterns of future data, now we build a classifier to estimate the label $y\in\mathbb{B}$.
We feed $\mathbf{p}$ into a FC network with a softmax for the output. Using \textit{Event Predictor} to denote the mapping $f_P = p(\hat{y}_{i,t}|\mathbf{p}_{i,t}; \boldsymbol{\omega}_P):\mathbb{R}^{Z} \xrightarrow{} \mathbb{R}^{\win_o}$, and $\boldsymbol{\omega}_P$ to denote the task-specific FC parameters,
we have
\begin{equation}\label{eq:Predictor}
    \begin{split}
    \hat{y}_{i,t} = \mathrm{Softmax}\left(\mathrm{FC}(\mathbf{p}_{i,t}; \boldsymbol{\omega}_P) \right)
    \end{split}
\end{equation}

\subsubsection{Masked Predictor Loss}
For a binary classifier, given true labels $y\in\{0,1\}$ and predicted probability $\hat{y}$, usually we use a cross-entropy loss 
\begin{equation}
   H({y}, \hat{y}) = -{y}\log(\hat{y}) -(1-{y})\log(1-\hat{y})
\end{equation}
For either the prediction of general or persistent hypoxemia, it's trivial and clinically less meaningful to predict the event when it already occurs, so we focus on predicting only the start of hypoxemia and leave the samples where hypoxemia already begins unlabeled (see Figure \ref{fig:labels}). To this end, a straightforward strategy is to directly exclude the unlabeled samples for both training and testing of the model, as in \cite{lundberg2018explainable,erion2017anesthesiologist}. However, the unlabeled samples can still provide our model useful information indicative of \spo tendency, and shape both Reconstructor $f_R$ and Forecaster $f_F$. 
Instead, we propose a masked loss for Predictor. For surgery $i$, we have the binary mask vector $\mathbf{m}_i=[m_{i,1}, m_{i,2}, ..., m_{i,T_i}]\in\mathbb{B}^{T_i}$ where $m_{i,t}=0$ indicates surgery $i$ at time $t$ is in hypoxemia, otherwise 1. The modified Predictor loss is:
\begin{equation}
    \mathcal{L}_P =\sum_i\sum_t H({m}_{i,t}{y}_{i,t}, {m}_{i,t} \hat{{y}}_{i,t} )
\end{equation}
In this way, the unlabeled data will be filtered out for Predictor but still used in training Reconstructor and Forecaster. This masked loss mechanism is very similar to semi-supervised learning where only part of samples are mapped to labels~\cite{liu2017semi}.

\subsubsection{Objective for Joint Learning}
For $i\in[1, N]$ and $t\in[1, T_i]$, we define the overall objective function to learn model parameters $\{\boldsymbol{\theta}_E, \boldsymbol{\omega}_R, \boldsymbol{\theta}_p, \boldsymbol{\omega}_F, \boldsymbol{\omega}_P\}$ for our end-to-end hybrid inference framework as minimizing the following joint loss:
\begin{equation}\label{eq:joint_loss}
    \begin{split}
     \mathcal{L} &= 
     \underbrace{\mathcal{L}_P}_\text{label decoder} + \ \ \  \lambda \underbrace{(\mathcal{L}_F + \mathcal{L}_R)}_\text{sequence decoders} \\
     &= \sum_i\sum_t \bigg[H\Big({m}_{i,t}{y}_{i,t}, f_P(\rep_{i,t};\boldsymbol{\theta}_p, \boldsymbol{\omega}_P)\cdot m_{i,t} \Big) \\
     &+ \lambda \bigg(\norm{\mathbf{u}_{i,t+L}- f_F(\rep_{i,t}; \boldsymbol{\theta}_p, \boldsymbol{\omega}_F)}^2 + \norm{\MTS_{i,t} -  f_R(\rep_{i,t}; \boldsymbol{\omega}_R)}_F^2 \bigg) \bigg]
    \end{split}
\end{equation}
where $\lambda$ is the weighing coefficient of both Reconstructor and Forecaster for simplicity.
The two sequence decoders $f_R$ and $f_F$ provide regularization to classifier $f_P$ with optimized data representation and future-indicative patterns. 
Given the end-to-end architecture, we can jointly update all the parameters during training. After the joint model is trained, we only need Predictor $f_P$ for label inference on new patients.

\section{Experiment}\label{sec:experiment}




\subsection{Data Preprocessing}
Examples with recorded \spo of less than 60\% were considered aberrant and excluded. We further excluded 5,606 cardiac surgery cases (with ICD-9 codes 745-747 and ICD-10 codes Q20-Q26) and 1,449 surgery cases with initial persistent low SpO2 (most likely cardiac related surgeries), in which \spo levels were likely affected by the surgical procedure. After extensive data cleaning, there are 72,018 surgeries and 18 channels of time series variables minutely sampled during surgical procedures. 
We aim to build a hypoxemia prediction system with 1-minute resolution.
However, not all variables were originally observed and recorded at every minute during surgery.
We use carry-forward imputation for the missing values in a gap between two observations less than 20 min, and fill those that are never observed or haven't been observed for the past 20 min with zeros. 
In addition, we concatenate a binary mask matrix with the time series input to indicate variable missingness.
All variables are standardized to zero mean and unit variance.

To further explore the effect of incorporating preoperative features (e.g., age, sex, weight) as part of the empirical analysis (see section \ref{sec:baseline} and \ref{sec:perf}), we also collected a set of preoperative variables for modeling. Table \ref{tab:feature} lists the 18 intraoperative variables and 9 preoperative variables used in this study.

\begin{table}[t]
\centering
\fontsize{8pt}{9.5pt}\selectfont
\caption{List of intraoperative and preoperative variables. }
\begin{tabular}{||ll||}
\hline
\textsc{Intraoperative Time Series}             & \textsc{Preoperative Variable} \\
\hline\hline
Invasive blood pressure, diastolic      & Age                          \\
Invasive blood pressure, mean           & Height                             \\
Invasive blood pressure, systolic       & Weight                            \\
Noninvasive blood pressure, diastolic   & Sex                       \\
Noninvasive blood pressure, mean        & ASA physical status                             \\
Noninvasive blood pressure, systolic    & ASA emergency status                             \\
Heart rate                              & Surgery type                             \\
SpO2                                    & Second hand smoke                             \\
Respiratory rate                        & Operating room                             \\
Positive end expiration pressure (PEEP) &                              \\
Peak respiratory pressure               &                              \\
Tidal volume                            &                              \\
Pulse                                   &                              \\
End tidal CO2 (ETCO2)                   &                              \\
O2 flow                                 &                              \\
N2O flow                                &                              \\
Air flow                                &                              \\
Temperature                             &                             \\
\hline
\end{tabular}
\label{tab:feature}
\end{table}

\subsection{Experimental Setup}
We randomly select 70\% of all the surgery cases for the model training, setting aside 10\% as a validation set for hyperparameter tuning and the other 20\% for model testing. We make sure that all data points from the same surgery case stay in the same subset of data.
We follow \cite{lundberg2018explainable} and set the prediction horizon as $\win_h=5$ min, given that it is short enough for a model to capture relevant and predictive information of potential future hypoxemia but long enough for clinicians to take actions. 

\subsubsection{Evaluation Metrics}
We use area under the receiver operating characteristic curve (ROC-AUC) and area under the precision-recall curve (PR-AUC) to evaluate the overall prediction performance averaged on all possible output threshold. Note that PR-AUC is more informative in evaluating imbalanced datasets~\cite{davis2006relationship}. We report the number of average false alarms per hours of surgery (False Ala./Hr) with a 5-minute \textit{redundant alarm suppression} window (i.e., consider as single alarm if a second alarm goes off within a 5-minute window), averaged on all possible model sensitivity. Please see more discussion on alarm suppression in \autoref{sec: practical}.

\subsubsection{Hyperparameters}
For both the two hypoxemia event prediction, we set the observation window $\win_o=20$ min.
For persistent hypoxemia prediction, we set the horizon of Forecaster $L=\win_h + 5=10$. For general hypoxemia prediction, we set $L=\win_h + 1=6$, based on the intuition that, given a 5 min prediction horizon, we need to see 10 min ahead to the future to speculate persistent hypoxemia and only 6 min for single \spo drops.

\setlength{\tabcolsep}{3.5pt}
\begin{table*}[t]
\centering
\fontsize{8.5pt}{10pt}\selectfont
\caption{Model overall performance on two types of hypoxemia condition.}

\begin{tabular}{||L{2.5cm}||C{1.3cm}C{1.3cm}C{1.8cm}|C{1.3cm}C{1.3cm}C{1.8cm}||}
\hline
\multirow{2}{*}{\textsc{Model}} & \multicolumn{3}{c|}{\textsc{Persistent Hypoxemia} ($\geq$ 5 min)} & \multicolumn{3}{c||}{\textsc{General Hypoxemia} ($\geq$ 1 min)} \\ 
                        & PR-AUC      & ROC-AUC         & False Ala./Hr    & PR-AUC    & ROC-AUC    & False Ala./Hr   \\ 
\hline\hline
LR        &  .0421  &  .9198  &  1.21  & .1213  & .8910  &  2.87\\
GBM   & .0570  & .9305  & .81  &  .1652  & .8932  & 1.44 \\
\hline\hline

LSTM        &   .0574  &  .9283  &  .69   & .1542  & .8920  &  1.78 \\
TCN \cite{bai2018empirical}  &  .0654  &  .9302  &  .51   &  .1811  &  .8956  &  1.32\\
FCN \cite{wang2017time}  & .0681  & .9354  &  .47  & .2005  &  .9024  &  1.05  \\
\hline\hline
LSTM-AE \cite{zhu2017deep} & .0695 &  .9321   &  .58  & .1772 & .8921  &  1.54\\
LSTM-CAE \cite{srivastava2015unsupervised} & .0734 &  .9345   &  .50  & .1823 & .8965  &  1.28 \\
TCN-AE  &   .0744  &  .9334  &  .49  &  .1842  &  .8971  &  1.25\\	
FCN-AE   &   .0801  &  .9440  &  .44  &  .2011  &  .9078  &  1.02\\
\hline\hline
\our-l   &  .0775  &  .9421  &  .44  &  .1897  &  .9011  &  1.16  \\
\our-t   &  \underline{.0866}  &  \underline{.9471}  &  \underline{.36}  &  \textbf{.2124}  &  \underline{.9167}  &  \textbf{.96}  \\
\our-f   &  \textbf{.0893}  &  \textbf{.9475}  &  \textbf{.34}  &  \underline{.2120} &  \textbf{.9196}  &  \underline{.98}\\
\hline\hline
GBM w/ PreOp \cite{lundberg2018explainable} &  .0716  &  .9322  &  .52  &  .1785  &  .8942  &  1.35\\
\our-f w/ PreOp &  \textbf{.1021}  &  \textbf{.9624}  &  \textbf{.28} &  \textbf{.2199}  &  \textbf{.9208}  &  \textbf{.95} \\
\hline
\end{tabular}

\label{tab:main}
\end{table*}

\subsection{Baseline Methods}
\label{sec:baseline}
We compare \ours to the following classical models, deep sequential models, and unsupervised pretraining based methods:

\begin{itemize}\setlength\itemsep{0em}
\item \textbf{LR}: Logistic Regression. Since LR cannot directly process time series, for fair comparison, we follow \cite{li2020deepalerts,fritz2019deep} to extract a series of summary statistics (e.g., min, max, trend, energy, kurtosis) that capture temporal patterns of history time series within the window of the same length as \ours.
\item \textbf{GBM}: Gradient Boosting Machines, employed by the state-of-the-art hypoxemia prediction system~\cite{lundberg2018explainable}, which is implemented using XGBoost~\cite{chen2016xgboost}. We use the same statistical features as in LR.

\item \textbf{LSTM}: Using stacked bi-LSTM for feature gathering and a FC block as the classifier, with layers configured the same way as the Event Predictor in \ours.
\item \textbf{TCN} \cite{bai2018empirical}: Temporal Convolutional Network with causal convolutions and exponentially increased dilation. It is configured the same way as the TCN module in \ours.
\item \textbf{FCN} \cite{wang2017time}: Full Convolutional Networks, a deep CNN architecture with Batch Normalization, shown to have outperformed multiple strong baselines on 44 benchmarks for time series classification.
\item \textbf{LSTM-AE}~\cite{zhu2017deep}: A deep LSTM classifier with the weights pretrained on an LSTM-AE.
\item \textbf{LSTM-CAE}~\cite{srivastava2015unsupervised}: A deep LSTM classifier with the weights pretrained on an LSTM-CAE that jointly reconstructs input and forecasts future input.
\item \textbf{TCN-AE} and \textbf{FCN-AE}: Replacing the RNN encoder in LSTM-AE \cite{zhu2017deep} with TCN and FCN for comparison.
\item \textbf{\our-l}, \textbf{\our-t} and \textbf{\our-f}: The \ours variants with the sequence encoder implemented by LSTM, TCN, and FCN.
\item \textbf{GBM w/ PreOp} and \textbf{\ours w/ PreOp}: For GBM w/ PreOp, the preoperative variables are added to the input of GBM as in \cite{lundberg2018explainable}. For \ours w/ PreOp, the preoperative features are directly concatenated with the data representation $\mathbf{p}_{i,t}$ in Eq. (\ref{eq:Predictor}).
\end{itemize}

\subsection{Implementation Details}\label{sec:setting}
For TCN and \our-t, we use 3 TCN blocks with number of filters set as 64 and dilation of each block set as 2, 4, and 8. For FCN and \our-f, we use 3 blocks for the FCN and use the filter sizes $\{64, 128, 64\}$ for each block.
Each FC block in \ours has only one hidden layer with rectified linear unit (ReLU) as the activation function. The number of neurons for each hidden layer in both LSTM and FC block, and the number of basis $M$ in the memory are all set as 128. We select the regularizer coefficient $\lambda$ from $\{10^{-4}, 10^{-3}, 10^{-2}, 10^{-1}, 10^{0}, 10^{1}\}$. The best $\lambda$ is 0.1 for persistent hypoxemia and 0.01 for general hypoxemia.

We use Adam as the optimizer with default learning rate and train the model with mini batches. For each batch, we feed 32 independent surgeries containing on average about 2,880 extracted examples into the model. We use the same settings for all deep baselines and all variants of \ours. All of them are trained for 50 epochs with early stopping and drop-out applied to prevent overfitting. The model with the lowest epoch-end classification loss at each run is saved and evaluated with test data.
The proposed model is implemented using TensorFlow 2.4.0 and Keras 2.4.3 with Python 3.8, and trained using NVIDIA GeForce RTX 3080 Ti GPUs and Intel Core i9-10850K 3.60GHz CPUs.

\section{Result and Discussion}\label{sec:result}

\subsection{Overall Performance}
\label{sec:perf}
\subsubsection{Overall performance}
\autoref{tab:main} summarizes the overall performance of all the models. Our proposed \ours framework outperforms all the baseline methods, achieving improved PR-AUC/ROC-AUC scores and \textbf{46\% and 32\% reduced average false alarm rate} over the state-of-the-art hypoxemia prediction system (GBM w/ PreOp) and the best deep baseline (LSTM-CAE), respectively. \ours effectively improves the efficacy of alerts.

\subsubsection{Feature Engineering vs. Representation Learning}
In general, the non-deep models LR and GBM
show inferior prediction performance compared to 
deep learning models. This may result from the less effective capacity of simple statistical features in capturing complex activity dynamics and temporality.
In contrast, armed with our proposed activity embedding, the deep models can learn deep representations capable of encoding more complicated patterns. 

\subsubsection{Supervised vs. Self-supervised Learning}
Among the deep models, those with AE-based pretraining (e.g., LSTM-AE, TCN-AE, FCN-AE) outperform their corresponding base model (e.g., LSTM, TCN, FCN). The self-supervised learning helps learn better data representations that benefit the discriminative task. Our \ours framework further achieve better performance by jointly learning better representations simultaneously optimized for prediction. The self-supervised component helps with contextual representation learning that potentially improves the robustness against extreme class imbalance.

\subsection{Ablation Study}
As shown in \autoref{tab:ablation}, we analyze the contribution of each component by removing it from \ours.
Note that for the variant w/o Memory, the memory layers are replaced with two stacked independent linear layers to maintain similar model complexity for fair comparison. We can see that for both the two prediction tasks (persistent \& general hypoxemia), each component in \ours plays essential role in improving the prediction performance.

\setlength{\tabcolsep}{3.5pt}
\begin{table}[t]
  \caption{Effect of each module in \ours.}
\centering
\fontsize{8.5pt}{11pt}\selectfont
\begin{tabular}{||lccc||}
\hline
  {\color[HTML]{000000} \textsc{Persist. Hypo.}} &
  {\color[HTML]{000000} PR-AUC} &
  {\color[HTML]{000000} ROC-AUC} &
  {\color[HTML]{000000} False Ala./Hr} \\ 
\hline\hline

  {\color[HTML]{000000} w/o Forecaster} &
  {\color[HTML]{000000} .0812} &
  {\color[HTML]{000000} .9416} &
  {\color[HTML]{000000} .42} \\ 

  {\color[HTML]{000000} w/o Reconstructor} &
  {\color[HTML]{000000} .0795} &
  {\color[HTML]{000000} .9367} &
  {\color[HTML]{000000} .45} \\

  {\color[HTML]{000000} w/o Memory} &
  {\color[HTML]{000000} .0765} &
  {\color[HTML]{000000} .9356} &
  {\color[HTML]{000000} .48} \\ 

\hline\hline
 {\color[HTML]{000000} \our-f} &  \textbf{.0893}  &  \textbf{.9475}  &  \textbf{.34} \\
\hline
\end{tabular}

\vspace{1em}

\begin{tabular}{||lccc||}
\hline
  {\color[HTML]{000000} \textsc{Gener. Hypo.}} &
  {\color[HTML]{000000} PR-AUC} &
  {\color[HTML]{000000} ROC-AUC} &
  {\color[HTML]{000000} False Ala./Hr} \\ 
\hline
\hline

  {\color[HTML]{000000} w/o Forecaster} &
    {\color[HTML]{000000} .2069} &
  {\color[HTML]{000000} .9108} &
  {\color[HTML]{000000} 1.01} \\

  {\color[HTML]{000000} w/o Reconstructor} &
  {\color[HTML]{000000} .1997} &
  {\color[HTML]{000000} .9022} &
  {\color[HTML]{000000} 1.14} \\ 

  {\color[HTML]{000000} w/o Memory} &
  {\color[HTML]{000000} .2007} &
  {\color[HTML]{000000} .9096} &
  {\color[HTML]{000000} 1.05} \\

\hline\hline
 {\color[HTML]{000000} \our-f} &  \textbf{.2120} &  \textbf{.9196}  &  \textbf{.98} \\
  
\hline
\end{tabular}\label{tab:ablation}
\end{table}

\subsection{Practical Effectiveness}\label{sec: practical}

\subsubsection{Alarm Suppression}
\begin{figure}[t]
  \centering
  \includegraphics[width=0.8\linewidth]{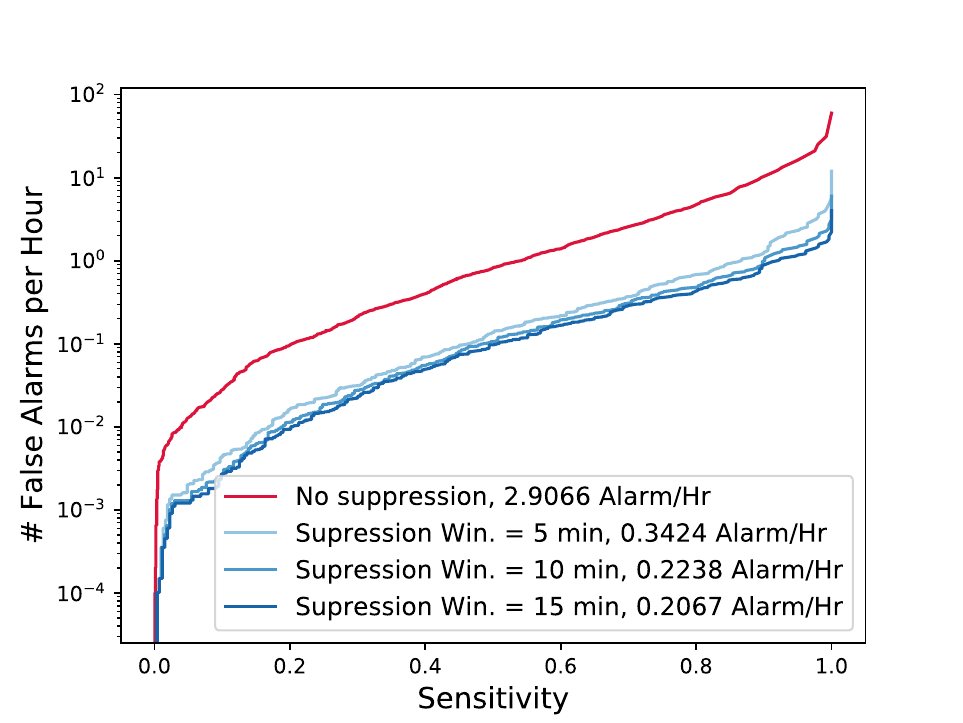}
  \captionof{figure}{Impact of alarm suppression window to false alarm rate (\our-f on persistent hypoxemia).}
  \label{fig:suppression}
\end{figure}

The \ours framework is designed to provide real-time prediction of near-term hypoxemia events at a one-minute resolution. For a continuous prediction system, multiple alarms raised by the prediction model within a short time window should be considered as the prediction of one approaching event, instead of multiple independent ones. To reduce alarm fatigue in a practical setting, we suppress redundant alarms within a short window (e.g., 5 minutes). Whenever there is any alarm going off within a certain time window to the first alarm, we silence the subsequent alarm and only consider the first alarm for true and false alarm evaluation. Figure \ref{fig:suppression} shows the impact of window size to the false alarm rates. We can see that by applying alarm suppression with a 5-minute window, the average false alarm rate dropped by 88\%. The alarm rate changes slightly when using much larger windows. Hence, we stick to 5 minutes as the window size for the remaining false alarm evaluation in this paper.

\subsubsection{False Alarm vs. Sensitivity}
To mitigate alarm fatigue, it is crucial for a clinical alarm system to minimize its false alarm rate given a sensitivity threshold. Hence, we plot the False Alarm vs. Sensitivity curve to evaluate the impact of our model on clinical practice, as shown in Figure \ref{fig:false}. 
In practical hypoxemia intervention and mitigation, the cost of predicting a false positive is much less than missing a true positive (i.e., relatively low cost of intervention vs. life-threatening persistent hypoxemia). Thus we prefer a high model sensitivity for hypoxemia prediction. 
For practical evaluation, given fixed sensitivity, Table \ref{tab:case} shows the false alarm rate comparison between GBM w/ PreOp and \our-f w/ PreOp. Our \ours framework is able to \textbf{reduce 64\% and 74\% false alarms} compared to the state-of-the-art system (GBM w/ PreOp \cite{lundberg2018explainable}) for sensitivity 0.8 and 0.6, respectively.

\begin{table}[t]
    \centering
    \fontsize{8pt}{11pt}\selectfont
    \caption{False alarm rate at sensitivity threshold 0.8 and 0.6.}
    \begin{tabular}{||c c c c||}
    \hline
\multirow{2}{*}{\textsc{Sensitivity}} & \multicolumn{2}{c}{\textsc{False Alarm} (Alarm/Hr)}  & \multirow{2}{*}{\textsc{Improv. (\%)}} \\ \cline{2-3}
 & \multicolumn{1}{c}{GBM w/ PreOp} & \our-f w/ PreOp &   \\ 
 \hline\hline
 0.8 & \multicolumn{1}{c}{0.89 } & 0.32    &   64\%\\ 
 0.6 & \multicolumn{1}{c}{0.31 } & 0.08  &  74\%\\
    \hline
    \end{tabular}
    \label{tab:case}
\end{table}

\begin{figure}[t]
  \centering
  \includegraphics[width=0.95\linewidth]{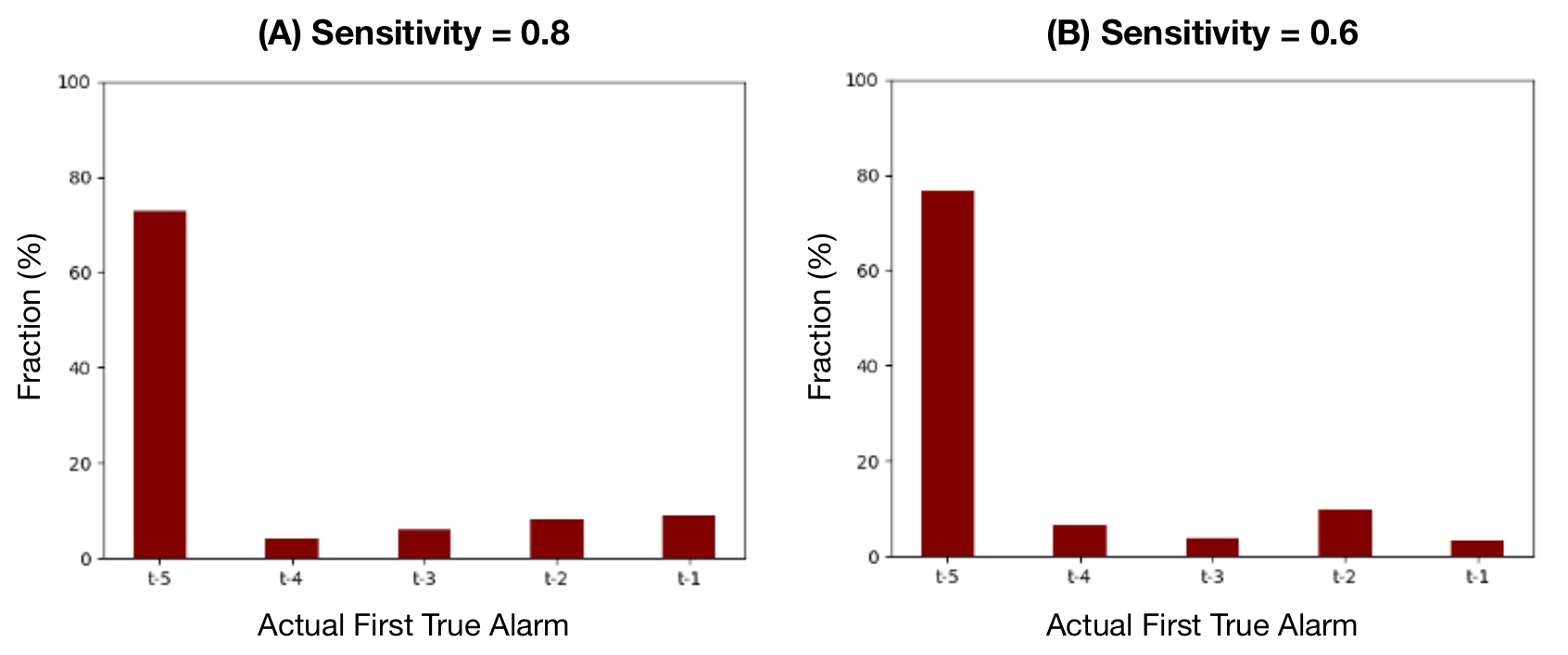}
  \captionof{figure}{Histogram of actual horizon of true alarms given sensitivity threshold 0.8 and 0.6, respectively (\our-f on persistent hypoxemia).}
  \label{fig:horizon}
\end{figure}

\begin{figure}[t]
  \centering
  \includegraphics[width=0.8\linewidth]{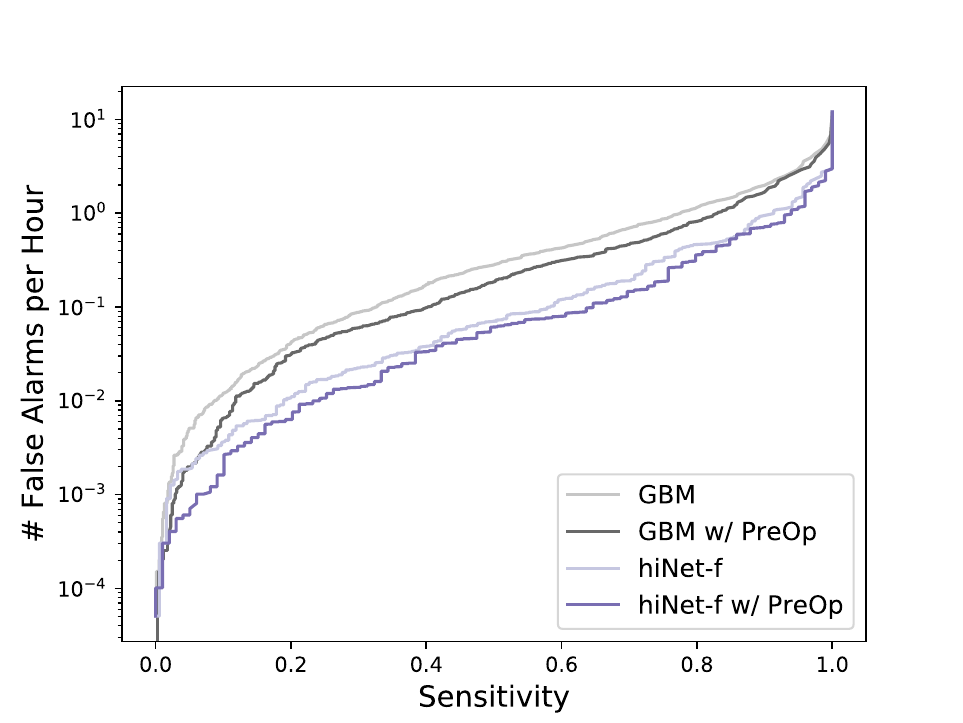}
  \captionof{figure}{False alarm rate vs. sensitivity for GBM and \our-f on persistent hypoxemia (5-minute suppression window).}
  \label{fig:false}
\end{figure}

\subsubsection{Lead Time of Alarms}
In the problem formulation (Section \ref{sec:formulation}) and label assignment for model training (Figure \ref{fig:labels}), we define the prediction horizon as 5 minutes and assign a positive label to each of the 5 minutes before the onset of the hypoxmeia event. Hence, the lead time of an alarm relative to the onset of the hypoxemia event may range from 1 to 5 minutes. For surgical care, it is important to analyze the \textit{actual} lead time of the alarms that can have a significant impact on clinical intervention during surgeries. Figure \ref{fig:horizon} shows during model inference the histogram of the actual lead time of the first true alarm for each persistent hypoxemia event in the testing set. We can see that for both the sensitivity threshold set as 0.8 and 0.6, our \ours predicted the event 5 minutes before it occurs for the majority of the persistent hypoxemia events, thus providing adequate lead time for intervention.

\subsection{Representation Learning}
\begin{figure}[t]
    \centering
    \includegraphics[width=0.99\linewidth]{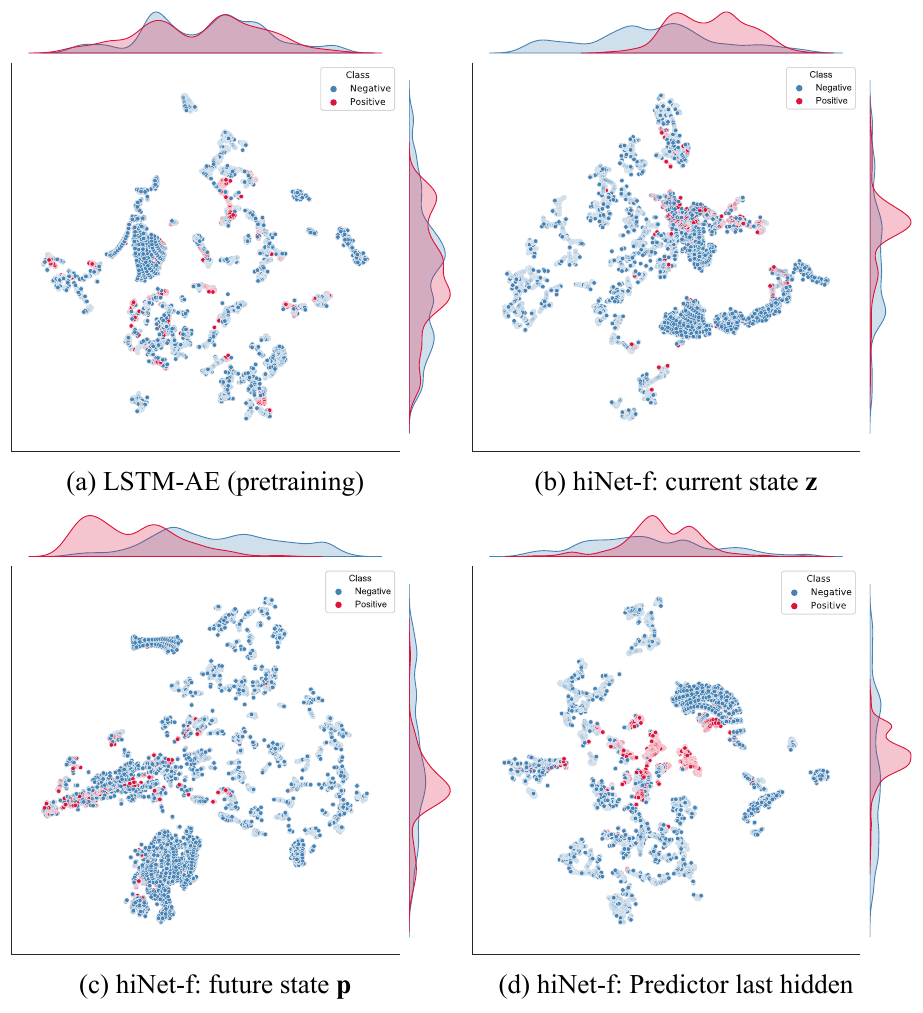}
    \caption{t-SNE visualization (perplexity = 80) and kernel density estimation (scaled) of learned representation by LSTM-AE and \ours on general hypoxemia prediction. (a) encoder output of LSTM-AE; (b) \ours current state $\mathbf{z}$; (c) \ours future state $\mathbf{p}$; and (d) output of last Predictor hidden layer in \ours.}
    \label{fig:tsne}
\end{figure}

As shown in \autoref{fig:tsne}, we extracted the latent representations learned by LSTM-AE and various layers of \ours for general hypoxemia prediction, and visualize these vectors in a 2D space using t-SNE~\cite{van2014accelerating}. Considering the extreme class imbalance, we randomly select 50 surgeries where persistent hypoxemia occurred and 50 hypoxemia-free surgeries from the test set for visualization purpose. As shown in Figure \ref{fig:tsne}, the representation of the same class in the latent space tends to group together in \ours, which enlarges the partitioning margin and makes them easier to classify. More explicit grouping patterns can be observed at layers closer to the Event Predictor output with stronger supervisory signal. In contrast, the representation learned by unsupervised LSTM-AE is structured but shows much less salient grouping patterns. We observe that, \ours is able to learn powerful task-specific representation via joint training, where supervisory signal is propagated to fine-tune latent representation towards task-specific effectiveness.

\subsection{Potential Limitations}
When producing labels for clinical outcomes such as hypoxemia, anaesthesiologist interventions can indirectly affect the prediction outcome~\cite{lundberg2018explainable}. As these interventions may affect certain vital parameters including \spo, models that use these parameters can learn when a doctor is likely to intervene and hence lower the risk of a potential high-risk patient. The ideal solution to this issue is to remove all samples where clinicians have intervened for model training. But this is difficult in practice, since fully identifying when clinicians are taking hypoxemia-preventing interventions is not possible. Hence, our model as all other previous learning based approaches~\cite{lundberg2018explainable,erion2017anesthesiologist} to this problem, must be based on the natural assumption that the model predicts hypoxaemia when clinicians are following standard procedures, including (possibly) taking interventions to prevent potential hypoxemia anticipated based on clinician's professional knowledge.

\section{Conclusion and Potential Impact}
Hypoxemia, especially persistent hypoxemia, is a rare but critical adverse surgical condition of high clinical significance. We developed hiNet, a novel end-to-end learning approach that employs a joint sequence autoencoder to predict hypoxemia during surgeries. In a large dataset of pediatric surgeries from a major academic medical center, hiNet achieved the best performance in predicting both general and persistent hypoxemia while outperforming strong baselines including the model used by the state-of-the-art hypoxemia prediction system. Our method produces low average false alarm rates, which helps mitigate alarm fatigue, an important concern in clinical care settings. 

This work has the potential to impact clinical practice by predicting clinically significant intraoperative hypoxemia and facilitating timely interventions. By tackling the challenging problem of predicting rare, but critical persistent hypoxemia, our model could help preventing adverse patient outcomes. 
We are currently working to implement our method directly into an application that can pull live intraoperative data streams from our health system’s EHR and present real-time predictions to surgeons and anesthesiologists in operating rooms. This will allow us to prospectively test the utility of our method in a real-world scenario by evaluating how accurately the alarms raises and how it is used on actual anesthetic interventions.

\section*{Acknowledgement}

This study was funded by the Fullgraf Foundation and the Washington University/BJC HealthCare Big Ideas Healthcare Innovation Award.


\begin{appendices}
\appendixpage

\section{Design Choice of Sequence Encoder}
\label{sec:design}

\subsection{Temporal Convolutional Networks}
Temporal convolutional networks (TCN) is a family of efficient 1-D convolutional sequence models where convolutions are computed across time \cite{bai2018empirical,lea2017temporal}. 
TCN differs from dypical 1-D CNN mainly by using a different convolution mechanism, dilated causal convolution. 
Formally, for a 1-D sequence input $\mathbf{X}=[\textbf{x}_1, ..., \textbf{x}_T]\in\mathbb{R}^{d\times T}$ and a convolution filter $\mathbf{f}\in\mathbb{R}^{k\times d}$, the dilated causal convolution operation $F$ on element $t$ of the sequence is defined as
\begin{equation}
    F(\mathbf{x}_t) = (\mathbf{X} *_d \mathbf{f})(t) = \sum_{i=0}^{k-1} \textbf{f}_i^T\cdot \mathbf{x}_{t-d\cdot i}, \ \ s.t., \ t\geq k, \textbf{x}_{\leq 0}:= 0
\end{equation}
where $d$ is the dilation factor, $k$ is the filter size, and $t - d\cdot i$ accounts the past. Dilated convolution, i.e., using a larger dilation factor $d$, enables an output at the top level to represent a wider range of inputs, effectively expanding the receptive field \cite{yu2015multi} of convolution. 
Causal convolution, i.e., at each step the convolution is only operated with previous steps, ensures that no future information is leaked to the past \cite{bai2018empirical}. This feature enables TCN to have similar directional structure as RNN models. 
Then the output sequence $\textbf{X}'\in\mathbb{R}^{k\times T}$ of the dilation convolution layer can be written as
\begin{equation}
    \textbf{X}' = [F(\mathbf{x}_1), F(\mathbf{x}_2), ..., F(\mathbf{x}_T)]
\end{equation}
Usually Layer Normalization or Batch Normalization regularization is applied after the convolutional layer for better performance \cite{lea2017temporal,bai2018empirical}. A TCN model is usually built with multiple causal convolutional layers with a wide receptive field that accounts for long sequences.

\subsection{Fully Convolutional Networks}
Full Convolutional Networks (FCN), a deep CNN architecturewith Batch Normalization, has shown compelling quality and efficiency for tasks on images such as semantic segmentation. 

An FCN model consists of several basic convolutional blocks. A basic block is a convolutional layer followed by a Batch Normalization layer and a ReLU activation layer, as follows:
\begin{equation}
\begin{split}
    \mathbf{y} & = \mathbf{W} * \textbf{x} + \textbf{d} \\
    \textbf{z} & = \text{BatchNorm}(\textbf{y}) \\
    \textbf{x}' & = \text{ReLU}(\textbf{z})
\end{split}
\end{equation}
where $*$ is the convolution operator. 

\end{appendices}


\bibliographystyle{acm}
\balance
\bibliography{reference.bib}

\end{document}